\documentclass[sigconf]{acmart}

\usepackage{algorithmic}
\usepackage{bm}
\usepackage{caption,subcaption}
\usepackage{graphicx}
\usepackage{xcolor}

\DeclareMathOperator*{\argmax}{\arg\!\max}

\setcopyright{acmcopyright}
\copyrightyear{2021}
\acmYear{2021}
\acmDOI{xx.xxxx/xxxxxxx.xxxxxxx}

\acmConference[DeMaL '21]{DeMaL '21:  International Workshop on Data-Efficient Machine Learning (DeMaL) at ACM Confierence on Knowledge Discovery and Data Mining (KDD) }{August 14--18, 2021}{Singapore}
\acmBooktitle{DeMaL '21:  International Workshop on Data-Efficient Machine Learning (DeMaL), August 14--18, 2021, Singapore}
\acmPrice{xx.xx}
\acmISBN{xxx-x-xxxx-XXXX-X/xx/xx}

\begin{document}

\title{Weakly Supervised Classification Using Group-Level Labels}

\author{Guruprasad Nayak}
\email{nayak013@umn.edu}
\affiliation{%
  \institution{University of Minnesota}
  \city{Minneapolis}
  \state{MN}
  \country{USA}
}

\author{Rahul Ghosh}
\email{ghosh128@umn.edu}
\affiliation{%
  \institution{University of Minnesota}
  \city{Minneapolis}
  \state{MN}
  \country{USA}
}

\author{Xiaowei Jia}
\email{xiaowei@pitt.edu}
\affiliation{%
  \institution{University of Pittsburgh}
  \city{Pittsburgh}
  \state{PA}
  \country{USA}
}

\author{Vipin Kumar}
\email{kumar001@umn.edu}
\affiliation{%
  \institution{University of Minnesota}
  \city{Minneapolis}
  \state{MN}
  \country{USA}
}

\renewcommand{\shortauthors}{Nayak et al.}

\begin{abstract}
In many applications, finding adequate labeled data to train predictive models is a major challenge. In this work, we propose methods to use group-level binary labels as weak supervision to train instance-level binary classification models. Aggregate labels are common in several domains where annotating on a group-level might be cheaper or might be the only way to provide annotated data without infringing on privacy. We model group-level labels as Class Conditional Noisy (CCN) labels for individual instances and use the noisy labels to regularize predictions of the model trained on the strongly-labeled instances. Our experiments on real-world application of land cover mapping shows the utility of the proposed method in leveraging group-level labels, both in the presence and absence of class imbalance.
\end{abstract}

\maketitle

\section{Introduction} \label{introSection}

Traditionally, classification models are trained using a set of training samples for which features and their corresponding class labels are known. However, in several practical applications, we are faced with two significant challenges, the first of which is the challenge of inadequate training data. The number of samples required to train an accurate classifier increases with (a) complexity of the chosen classifier and (b) the heterogeneity of the feature space. For example, consider the example of using satellite images to map the extent of urban areas on the surface of the Earth. Urban areas (and non-urban areas) are heterogeneous enough globally that they require complex models like deep neural networks to predict them accurately. This creates the need for large number of training samples which is hard to meet in this application since manual annotation is the primary source of obtaining labeled data. Similar challenges occur in several other application areas of predictive modeling where the scale and variety of data instances is large.

The second key challenge that we address in this paper is that of class imbalance. In many applications, it is common to have an imbalance in the number of instances belonging to the two classes like in the case of spam identification \cite{bingLiu} or land cover monitoring problems like urban areas \cite{schneider} or burned forests \cite{RAPT}. Thus, the learning algorithms have to optimize for the performance on the rare class unlike most standard classifiers that optimize for measures like accuracy.

While each of these two challenges are significant by themselves, the problem becomes much harder when they occur simultaneously, as is the case in many practical applications such as land cover mapping. In the case of inadequate training data, the number of instances from the rare class might be too few or the class imbalance might be misrepresented owing to limited samples.

\begin{figure}
    \centering
    \includegraphics[width = 8cm]{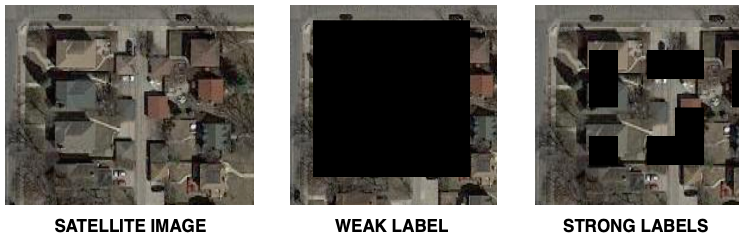}
    \caption{Example of group-level labels for the application of urban area mapping. The group-labels are obtained from a manual annotator who marks a general region as having buildings, which creates the group-level labels, as opposed to marking actual boundaries around each building, which would constitute the actual instance-level labels. Note that obtaining group-level labels is less expensive than instance-level labels.}\label{groupLevelLabelsEx}
\end{figure}

One way to handle the challenge of not having adequate training samples is to use weak supervision. In a weakly-supervised learning \cite{zhouWLIntro} scenario, you have very few (if any) training samples that have exact labels corresponding to the target variable. However, you have plenty of weakly-labeled instances i.e you have an imperfect version of the target variable for these instances. The idea is that, by modeling the imperfection in the weak labels, we can mitigate the lack of (strongly-labeled) training data. In many binary classification problems, weak supervision might be available in the form of group-level labels. Aggregate labels are common in several domains where annotating on a group-level might be cheaper or might be the only way to provide annotated data without infringing on privacy. Figure \ref{groupLevelLabelsEx} shows an example of group-labels in the application of mapping urban areas using satellite data. It is far less expensive to annotate that a general region has houses and hence is urban than to individually mark a polygon around each house. Another example of cheap group-level labels, in the context of land cover mapping is when you are trying to classify instances at a finer resolution (e.g, 10m *10m pixels), but you have weak labels available at coarser resolution (e.g, 250m*250m pixels). This happens because most existing work has been done on creating coarser resolution maps and hence many accurate products exist at this resolution. Thus, each of the 625 10m*10m pixels within a 250m*250m pixel become part of the same group, for which a single binary label is available.  A good example of when privacy concern leads to group-level labeling is census data when information is aggregated at the level of administrative boundaries. Group-level labels are common in several other domains including drug activity prediction \cite{MILdrug}, object recognition \cite{kuck}, image retrieval \cite{MILir} and text categorization \cite{MILtext}. Note that in our case, the group-level labels are binary as well. However, distribution of instance labels within a group and the size of the group are both variable.

Based on prior research, there can be two approaches to using group-level binary labels for instance-level classification. The first is to treat it as  multiple instance learning problem \cite{MILsurvey1} where the goal is to make group-level predictions. An aggregation function (eg. max or mean) is assumed to generate group-level predictions from predictions on constituting instances and the classifier is optimized for group-level performance metrics. This approach is limiting since it does not optimize for instance-level performance. The second approach is to propagate the group-level label to all instances within the group and treat this as a noisy label for learning an instance-level classifier. Most approaches for handling label noise rely on the Class Conditional Noise (CCN) assumption. Class Conditional Noise (CCN) is a well-studied model for learning with label noise \cite{menon, RAPT, raykar, bingLiu}. Class conditional independence is also a key assumption in several prior papers on using group-level supervision with label proportions \cite{quad, patrini}. In the group-label setting, this assumption would mean that the class-conditional distribution of the instances is independent of the groups. However, this assumption may not completely hold in practice. For example, in figure 1, there are certain kinds of non-urban areas like farms that are much more likely to be adjacent to urban areas and hence to be a part of urban-labeled groups than otherwise.

In this paper, we consider the scenario where we have very few strongly labeled samples and a large number of weakly labeled instances for training. The strongly-labeled instances are the regular training set used for learning classification models, with features and labels for individual instances. The weakly-labeled instances have instance-level features, but labels are only available on a group-level. We propose to train a regular classifier using the strongly-labeled instances and regularize it using its estimated performance on the weakly-labeled instances through CCN modeling. While the CCN modeling of group-labels for constituting instances may not exactly hold in practice and using it by itself might be limiting, as explained previously, using it together with strong supervision can help mitigate this limitation. The strongly-labeled instances are free of the label noise and hence have no bearing on the CCN modeling of the weakly-labeled instances, while the weakly-labeled instances provide the benefit of being available in plenty which the strong supervision lacks. Our approach leverages this complementarity between strong and weak supervision to train robust and accurate models.

\subsection{Contributions}
In a nutshell, the contributions of this work are as follows:
\begin{enumerate}
  \item We formalize the commonly-occuring weak label learning scenario of binary group-level labels in the context of instance level binary classification.
  \item We propose a principled approach to use group-level labeled instances (in addition to instance-level labeled instances) to train classification model to optimize for instance-level classifier performance. By modeling group-labels as CCN labels for constituting instances, we leverage prior research to optimize instance-level metrics.
  \item The proposed approach can also optimize for class imbalance, if it is known to exist in the data. This is especially relevant in the context of inadequate training data when the limited number of strongly-labeled instances may misrepresent the class imbalance or might not have enough samples from the rare class.
  \item Our method demonstrably leverages the group-level labels for landcover classification tasks (urban areas, burned vegetation) which are problems of great environmental and societal significance.
\end{enumerate}

\section{Related Work}

\subsection{Learning with group-level supervision}
Group-level supervision has been extensively studied before, most prominently as the paradigm of multiple instance learning (MIL), first introduced in \cite{MILdrug}. In a MIL problem setting, the goal is to learn a classifier to predict the label for groups of instances. The classic MIL assumption \cite{MILsurvey1} is that the binary label of the group is the disjunction of the unknown binary labels of its constituting instances i.e the presence of a positive label instance within a group makes the group label positive. The classic MIL assumption has been extended to more general threshold settings like in \cite{tao2004svm}. Alternately, the MIL problem can be solved using single instance classification algorithms by learning a embedding function for groups like in \cite{MILembed}. However, unlike the classic MIL setting, our goal is to predict the instance-level labels, not the group-label.

More recently, group-level supervision, in the form of label proportions, has been studied for learning instance-level classifiers. In this setting, the proportion of constituent instances belonging to the each class is known for each training group. There are two kinds of approaches in this category - probabilistic and non-probabilistic. The probabilistic models \cite{quad, patrini} assume that the class-conditional distribution of the data is independent of the bags. Using this assumption, they estimate a sufficient statistic from the label proportions on the training groups and use it to maximize the likelihood for instance-level prediction. The non-probabilistic approach \cite{kuck, labelProportions} is similar to the threshold-based MIL setting where the underlying instance labels are generated to remain consistent with the label proportions known for the corresponding groups. While we borrow the class conditional independence modeling assumption from these label proportion methods, our method does not assume the knowledge of proportion of labels on training groups, which might be harder to come by in many applications such as land cover mapping, where getting binary group labels might be easier. Moreover, to the best of our knowledge, none of these methods do rare class optimization on instance-level prediction.

\subsection{Learning with noisy labels}
Learning with noisy labels is a problem of great practical significance and has been widely studied in literature \cite{labelNoiseSurvey, blum}. A large number of papers try to handle this problem by assuming a class conditional noise (CCN) model, where the probability of corrupting the label is the same for all instances within a given class. Under the CCN assumption, some approaches like \cite{natarajan, patriniNoise} try to estimate a surrogate loss which is corrected for the noise in the labels. Other CCN based approaches try to estimate the noise rate before optimizing the desired metric on the noisy data \cite{menon, RAPT}. This approach has also been used in crowdsourcing settings \cite{raykar} where we have multiple sources of noisy labels, each with its own noise rate. Recently, \cite{goldbergerCCNLayer} proposed to augment deep neural networks with a noise adaptation layer that learns the noise rates along with the neural network parameters in an EM-style framework. The second category of approaches in learning with noisy labels train the classifier, usually a neural network, to learn from clean information initially before subjecting it to noisy labels. Some of these methods \cite{li2017learning, veit2017learning} require access to a clean data set to pre-train the neural network while other more recent methods like \cite{coteaching} use the memorization effect \cite{memorization} of deep learning models where these models are known to learn the clean information in initial iterations before overfitting on the noisy labels. Again, most methods in the noisy label category, do not optimize for rarity, except for \cite{RAPT}. Also, to the best of our knowledge, these methods have not been applied to the group-label weak supervision problem.

\section{Method} \label{methodsSection}
\subsection{Problem setting}
We consider a two-class classification problem where the goal is to learn a model that predicts the binary label $y \in \{0,1 \}$ for a test instance given its features $\bm{x}$. During the training phase, we have two sets of instances:
\begin{itemize}
    \item [(a)] \textbf{Strongly-labeled} data set of $N_s$ instances $\{ (\bm{x_i}, y_i) \}_{i=1}^{N_s}$ having features $\bm{x_i}$ and labels $y_i$ for each instance $i$.
    \item [(b)] \textbf{Weakly-labeled} data set of $N_g$ instances $\{ (\bm{x_i}, g_i) \}_{i=1}^{N_g}$ having features $\bm{x_i}$ and labels $g_i \in \{0,1\}$ of the corresponding group for each instance $i$.
\end{itemize}
Note that the distribution of true instance labels within a group and the size of the group are both variable. Also note that due to the relatively high cost involved in labeling individual instances, it is more likely that $N_s << N_g$.

\begin{figure}
    \centering
    \includegraphics[scale = 0.5]{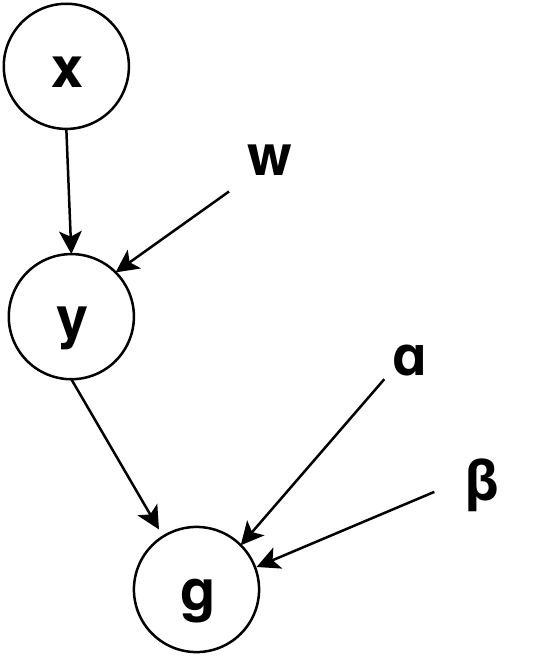}
    \caption{Graphical model diagram for Class Conditional Noisy (CCN) labels $g$, as they relate to features $\bm{x}$ and true labels $y$. The model assumes $g \perp \! \! \! \perp \bm{x} | y$. The conditional probability $g|y$ is encoded by parameters $\alpha$ and $\beta$, while parameter $w$ defines the distribution $y|\bm{x}$.}\label{CCNGraphicalModel}
    % \vspace{-1cm}
\end{figure}

\subsection{Modeling group-labels as CCN labels} \label{CCNmodel}
Class Conditional Noise (CCN) is a widely studied model for noisy labels \cite{menon}, \cite{natarajan}, \cite{blum}. Figure \ref{CCNGraphicalModel} shows a graphical model for the CCN model. The key assumption of the CCN model is $g \perp \! \! \! \perp \bm{x} | y$, where $g, y, \bm{x}$ denote the corrupted label, true label and the feature vector respectively. In other words, given a particular class, the probability of the noisy label making an error is the same for all instances within the class. For a two-class problem, the noise rates can be encoded as,
\begin{equation*}
    \begin{split}
    \alpha &= Pr(g = 0 | y =1) \\
    \beta &= Pr(g =1 | y = 0)
     \end{split}
\end{equation*}

\subsubsection{Assumptions for using group-level labels as weak labels for instance-level classification}
Group-level labels can be used to learn predictive models on an instance-level under two assumptions. \\
%Here, we discuss how those assumptions translate to the group-level labels from which our CCN weak labels are derived as explained in section \ref{pseudolabels}.
\textbf{(a) Instance-level pseudo-labels are CCN} CCN assumption assumes that the probability of flipping the class label on an instance is the same for all instances in the class irrespective of their feature values. Since we are propagating the group-label to all instances within the group, the 0-labeled instances within a 1-label group are being flipped to 1 (and vice-versa). For the CCN assumption to hold, these flipped instances should be no different in the feature space than instances that don't get flipped i.e 0-label instances within 0-label bags. In other words, the 0-label instances within 1-label bags and those within 0-label bags come from the same distribution (and vice-versa for 1-label instances). \\
\textbf{(b). Sum of noise rates is less than 1 \cite{blum, natarajan}}
This assumption states i.e $\alpha + \beta < 1$,
which can also be written as $((1-\alpha) + (1- \beta))/2 > 0.5$.
$1-\alpha$ is the probability that the true label is 0 when the CCN label is 0 (negative class accuracy) and $1-\beta$ is the probability that the true label is 1 when the CCN label is 1 (positive class accuracy). So, intuitively, this assumption is saying that the probability of the CCN label being true is better than 0.5 i.e the probability that a noisy label generated at random would be correct i.e. the CCN label itself is a weak learner.
    \begin{equation*}
        \begin{split}
            &\alpha + \beta < 1  \\
          &\implies Pr(y_g = 1 | y = 0) + Pr(y_g = 0 | y = 1) < 1 \\
            &\implies \frac{\sum_{y_i = 1 \& y_{g_i} = 0} 1}{\sum_{y_{i} = 0} 1} + \frac{\sum_{y_i = 0 \& y_{g_i} = 1} 1}{\sum_{y_{i} = 1} 1} < 1 \\
        \end{split}
    \end{equation*}
i.e the sum of the fraction of 1-label instances that fall in 0-bags and the fraction of 0-label instances that fall in 1-bags should be less than 1.

\subsection{Training objective}
The key idea of our weakly-supervised classification algorithm is to regularize the performance of a classifier trained on the strongly-labeled training set with its performance on the weakly-labeled instances. For the classification model, we consider a differentiable function $f(\bm{x})$ that models $Pr(y =1 | \bm{x})$ and a threshold $\gamma$ that converts it to a binary prediction $\hat{y}$ i.e $\hat{y_i} = 1$ if $f(\bm{x}) > \gamma$ and $0$ otherwise. Because the step-function is not differentiable at the threshold, we use a soft approximation to the step function, that thresholds the numeric value $f(\bm{x})$ at $\gamma$ as shown below
\begin{equation}
\hat{y^{\gamma}} = \frac{1}{1 + e^{-s*(f(\bm{x}) - \gamma)}}
\end{equation}
where $s$ is a large constant that controls how quickly $\hat{y_{\gamma}}$ goes to 0/1 on either end of the threshold $\gamma$. \\
The training objective is a sum of the objective over the two kinds of training instances - strong ($O_s$) and weak ($O_g$). The optimal parameters $\hat{w}$ for $f$ are estimated as,

\begin{equation}
    \label{genObjFunc}
    \begin{split}
        \hat{w} &= \argmax_{w} \max_{\gamma} O(w;  \gamma) \\
        &= \argmax_{w} \displaystyle\left[ \max_{\gamma} O_s(w; \gamma) + \lambda O_g(w; \gamma) \displaystyle\right]
    \end{split}
\end{equation}
where $\lambda$ is a weighting parameter fixed using cross-validation. We use gradient descent to optimize the above objective function. To keep the objective function differentiable, the maximum over different values of $\gamma$ is computed using a softmax function.

\subsubsection{$O_s$: Objective on strongly-labeled instances}
The performance of model $f$ with parameters $w$ for a given threshold $\gamma$ over the strongly-labeled samples $O_s(w; \gamma)$ is simply the classification accuracy, which can be computed as,
\begin{equation} \label{eqOs}
O_s(w; \gamma) = (1/N_s)*\sum_{i=1}^{N_s} \hat{y^{\gamma}_i} y_i + (1-\hat{y^{\gamma}_i})(1-y_i)
\end{equation}

\subsubsection{$O_g$: Objective on weakly-labeled instances}
Prior research has shown that it is possible to optimize certain performance metrics using surrogate losses that require only CCN labels (not the true labels). We use the surrogate versions of metrics proposed in these works to define the objective on weakly-labeled instances. We consider 2 cases, depending on the presence or absence of class imbalance.
\begin{enumerate}
    \item {\emph{Balanced case: } In \cite{menon}, the authors showed that the accuracy of a classifier computed using CCN labels can be used as a surrogate for its true accuracy. Thus, for the balanced class scenario, we define our objective on weakly-labeled instances to be the surrogate accuracy i.e
    \begin{equation} \label{eqOwbl}
        O_g(w; \gamma) = (1/N_g)*\sum_{i=1}^{N_g} \hat{y^{\gamma}_i} g_i + (1-\hat{y^{\gamma}_i})(1-g_i)
    \end{equation}}
    \item {\emph{Imbalanced case: } Similarly, for rare class scenarios, where metrics like precision and recall are more important, the authors in \cite{RAPT} showed that the geometric mean of precision and recall (G-measure) can be optimized using a surrogate, computed using only the knowledge of CCN labels. This surrogate for G-measure can be used as our objective on weakly-labeled instances like so,
    \begin{equation} \label{eqOwImbl}
        \begin{split}
            O_g(\gamma) &= (Pr(g = 1 | \hat{y} =1) - \beta)^2{Pr(\hat{y} = 1)} \\
            &= ( \frac{\sum_{i=1}^{N_g}\hat{y^{\gamma}_i} g_i/N_g}{\sum_{i=1}^{N_g} \hat{y^{\gamma}_i}/N_g} - \beta)^2 \sum_{i=1}^{N_g} \hat{y^{\gamma}_i}/N_g \\
            &= \frac{(\sum_{i=1}^{N_g}\hat{y^{\gamma}_i} g_i - \beta\sum_{i=1}^{N_g} \hat{y^{\gamma}_i})^2}{N_g\sum_{i=1}^{N_g} \hat{y^{\gamma}_i}}
        \end{split}
    \end{equation}
    where $\beta = Pr(g = 1 | y =0)$ is a property of the weak label that can be estimated using the strongly labeled samples.}
\end{enumerate}
\emph{Estimating } $\beta$: To estimate $\beta = Pr(g = 1 | y =0)$, we follow the same procedure as the authors in \cite{RAPT}. Ideally, one would want sufficient number of instances with both $y$ and $g$ labels. However, since this is not the case, we use the strongly-labeled samples to estimate sufficient samples with $y =0$. We learn a classifier with just the strongly-labeled samples. Even if this classifier is not very accurate, we expect the instances with very low probability assigned from this classifier to be mostly negative. Thus, we take the bottom 5\% of weakly-labeled instances with lowest values of estimated $Pr(y =0|x)$ using this classifier trained with only strongly-labeled instances. The estimate for $\beta$ is the fraction of these instances with $g=1$.

\section{Evaluation} \label{evaluationSection}
We evaluate the methods proposed in this paper on 7 data sets from two real world applications - urban area detection and burned area detection, that use satellite-collected observations of different locations to automatically track changes changes on the surface of the Earth. Group-level labels are derived from auxiliary satellite products, that are easily available.  Since many of these satellite products are available at a coarser resolution in comparison to the locations being classified, they act as group-level labels for the classifier being trained. For the urban extent mapping application, we consider group-level label derived from the Night time light satellite product \cite{lpdaac}. The average group size in this case is about 2000. Similarly, for the burned area mapping problem, we use another satellite signal called Active Fire \cite{lpdaac}. The average group size in this case is about $9$. More details about these data sets are provided in table \ref{tableDataSetSummary}. The satellite signals used in the paper are freely available from \cite{lpdaac} and will also be made available through the author's personal website after publication, along with the code for the algorithms proposed in this paper.

\begin{table*}
    \centering
    \caption{Summary of data sets used in this paper.}
    \begin{tabular}{|c|c|p{1.7cm}|p{2cm}|p{2cm}|c|c|p{2cm}|}
    \hline
    \textbf{Data set}&\textbf{Phenomenon}&\textbf{Strongly labeled instances}& \textbf{Average group size}&\textbf{Weakly labeled instances}& \textbf{$\alpha$ }& \textbf{$\beta$ }& \textbf{skew = \#neg/\#pos}\\ \hline \hline
    D1 & Fires, CA & 200 & 8.91 & 20535 & 0.338 & 0.064 & 1 \\ \hline
    D2 & Fires, MT & 200  & 8.94 & 20506 & 0.309 & 0.0057 & 1 \\ \hline
    D3 & Fires, GA & 100  & 8.82 & 582 & 0.888 & 0.04 & 1 \\ \hline
    D4 & Urbanization, MN & 20  & 2034.67 & 11190685 & 0.067 & 0.0039 & 1 \\ \hline
    D5 & Fires, CA & 200  & 8.91 & 17482 & 0.487 & 0.029 & 4.81 \\ \hline
    D6 & Fires, MT & 200  & 8.94 & 17428 & 0.435 & 0.028 & 4.80 \\ \hline
    D7 & Fires, GA & 100  & 8.82 & 977 & 0.898 & 0.051 & 2.82 \\ \hline
    \end{tabular}
    \label{tableDataSetSummary}
\end{table*}

\subsection{Comparison with baselines} \label{baselineComp}
For our experiments, we compare the performance of our method (called \emph{WeaSL}) against the following baselines:
\begin{itemize}
    \item[(a)] \textbf{OnlyStrong} Only use the strongly labeled data and no weak supervision to learn the model parameters.
    \item[(b)] \textbf{MIL (balanced) \cite{MILR}} MI/LR is a multi-instance learning algroithm proposed in \cite{MILR}, where the goal is to predict labels for groups of instances. The algorithm learns an instance level classifier and uses a mean-aggregation on the prediction of instances within a bag to predict the group-label. We learn MIL on groups of instances in the weakly-labeled data set and use the learned instance-level classifier within MIL to make predictions on the test set. We also experimented with max-aggregation for generating the group-label from instance labels, however, mean-aggregation seemed to work better for our data sets.
    \item[(c)] \textbf{MIL (imbalanced)} This is similar to MIL (balanced) \cite{MILR}, however, instead of optimizing for accuracy on the group-level predictions, this method optimizes for F-measure. Note that optimizing for F-measure on the group-predictions is not equivalent to optimizing F-measure on the instance level. In an extreme situation, the group-level labels might be balanced while the instances themselves can be skewed, as in the group-purity experiment in the supplement.
    \item[(d)] \textbf{OnlyWeak \cite{menon, RAPT}} Variant of WeaSL where we ignore the strongly-labeled instances and only optimize the performance (accuracy or G-measure) over weakly-labeled instances. Thus, instead of considering $O_s+ \lambda O_w$ in equation \ref{genObjFunc}, our objective function is simply $O_w$ in this case. This method is similar to the learning algorithms used to learn with CCN labels in \cite{menon, RAPT}.
    \item[(e)] \textbf{Co-teaching \cite{coteaching}} This is another approach of handling noisy labels that relies on the memorization effects of the classification models instead of assuming a CCN model for label noise. This is also a method that uses only weak supervision and no strong supervision.
\end{itemize}

Unless specified otherwise, the base classifier for all methods in this section is a neural network with 2 hidden layers with 128 and 64 units respectively.
\subsubsection{Evaluation metrics:} We evaluate the performance on the balanced data sets using accuracy and on the imbalanced data sets using F-measure.

\subsubsection{Balanced data sets:} Table \ref{tableBalancedResults} reports the average accuracy over 5 iterations for the balanced data sets in table \ref{tableDataSetSummary}. Since we have very few strongly labeled samples, the \emph{OnlyStrong} baseline does quite poorly, unlike the other methods that leverage weakly labeled data. Methods  \emph{MIL (balanced)} and \emph{OnlyWeak} ignore the strongly-labeled instances and use only weakly-labeled instances for training. \emph{MIL (balanced)} is designed to predict group-labels and thus, does not optimize for instance-level prediction performance. In comparison, \emph{OnlyWeak} optimizes for instance-level accuracy through CCN modeling of the group-label for individual instances. Thus, it mostly seems to do better than {MIL (balanced)}. \emph{Co-teaching} is another method that uses only weak supervision through a different noise handling strategy. Thus, it performs similar to \emph{OnlyWeak}. However, since the noise modeling assumption may not exactly hold in the data, these methods fall short of \emph{WeaSL} that benefits from using the weakly-labeled data in conjunction with the strongly-labeled data.

\begin{table}[h]
    \vspace{-0.3cm}
    \centering
    \caption{Comparison with baselines: balanced data sets (Accuracy)}
    \begin{tabular}{|c|c|c|c|c|}
        \hline
        \textbf{Method} & \textbf{D1} & \textbf{D2} & \textbf{D3} & \textbf{D4} \\ \hline \hline
        OnlyStrong & $\underset{(0.013)}{0.741}$ & $\underset{(0.011)}{0.754}$ & $\underset{(0.022)}{0.688}$ & $\underset{(0.075)}{0.681}$ \\ \hline
        OnlyWeak & $\underset{(0.004)}{0.763}$ & $\underset{(0.002)}{0.769}$ & $\underset{(0.022)}{0.529}$ & $\underset{(0.009)}{0.791}$ \\ \hline
        WeaSL & $\underset{(0.005)}{\textbf{0.819}}$ & $\underset{(0.003)}{\textbf{0.805}}$ & $\underset{(0.008)}{\textbf{0.714}}$ &  $\underset{(0.004)}{\textbf{0.847}}$ \\ \hline
        MIL (balanced) & $\underset{(0.035)}{0.632}$ & $\underset{(0.013)}{0.655}$ & $\underset{(0.001)}{0.554}$ & $\underset{(0.005)}{0.714}$ \\ \hline
        Co-teaching & $\underset{(0.002)}{0.757}$ & $\underset{(0.001)}{0.763}$ & $\underset{(0.031)}{0.511}$ & $\underset{(0.001)}{0.803}$ \\ \hline
    \end{tabular}
    \label{tableBalancedResults}
    \vspace{-0.3cm}
\end{table}

\subsubsection{Imbalanced data sets:} Table \ref{tableImbalancedResults} reports the average F-measure over 5 iterations for the imbalanced data sets in table \ref{tableDataSetSummary}. As shown in table \ref{tableDataSetSummary}, note that unlike the balanced case where the number of instances in both classes were the same, here, the majority class has more than $90\%$ of the instances in some cases. Similar to the balanced data sets, the performance of \emph{OnlyStrong} is relatively poor because it only uses limited strongly-labeled training samples. \emph{MIL (balanced)} and \emph{MIL(imbalanced)} optimize for group-level prediction performance and hence they do not perform as well as our method that explicitly optimizes for instance-level precision and recall. Note that \emph{MIL (imbalanced)} still optimizes F-measure on the group-level and thus, performs sub-optimally on instance-level predictions. Similar to \emph{MIL}, \emph{OnlyWeak} and \emph{Co-teaching} consider only weak supervision for training. But instead of optimizing group-level performance, \emph{OnlyWeak} models the group-labels as CCN labels for instances and optimizes instance-level G-measure. Thus, its performance is significantly better than \emph{MIL}. However, it still falls short of \emph{WeaSL} that not only optimizes for rarity but also uses strongly-labeled instances that helps it to overcome the limitations of CCN modeling.

\begin{table}[h]
    \centering
    \caption{Comparison with baselines: imbalanced data sets (F-measure)}
    \begin{tabular}{|c|c|c|c|}
        \hline
        \textbf{Method} & \textbf{D5} & \textbf{D6} & \textbf{D7} \\ \hline \hline
        OnlyStrong & $\underset{(0.027)}{0.462}$ & $\underset{(0.0187)}{0.4918}$ & $\underset{(0.044)}{0.5947}$  \\ \hline
        OnlyWeak & $\underset{(0.008)}{0.461}$ & $\underset{(0.008)}{0.437}$ & $\underset{(0.025)}{0.429}$  \\ \hline
        WeaSL & $\underset{(0.002)}{\textbf{0.602}}$ & $\underset{(0.012)}{\textbf{0.629}}$ & $\underset{(0.003)}{\textbf{0.623}}$  \\ \hline
        %SSRManifold & 0.3017 & 0.2645 & 0.5443  \\ \hline
        MIL (balanced) & $\underset{(0.044)}{0.294}$ & $\underset{(0.043)}{0.334}$ & $\underset{(0.056)}{0.215}$  \\ \hline
        MIL (imbalanced) & $\underset{(0.016)}{0.414}$ & $\underset{(0.021)}{0.454}$ & $\underset{(0.048)}{0.355}$  \\ \hline
        Co-teaching  & $\underset{(0.003)}{0.560}$  &   $\underset{(0.004)}{0.572}$ &  $\underset{(0.036)}{0.225}$\\ \hline
    \end{tabular}
    \label{tableImbalancedResults}
\end{table}

Table \ref{tableBetaEst} reports the estimated $\beta$ along with its true value for the 3 dats sets. As we can clearly see, the estimates are close enough to the actual values.
\begin{table}[h]
    \centering
    \caption{Estimates of noise rate $\beta = Pr(g=1|y=0)$}
    \begin{tabular}{|c|c|c|c|}
        \hline
        & D5 & D6 & D7 \\ \hline
        $\beta$ (actual) & 0.029 & 0.028 & 0.051\\ \hline
        $\beta$ (estimated) & 0.027 & 0.035 & 0.041\\ \hline
    \end{tabular}
    \label{tableBetaEst}
\end{table}

\subsection{Further analysis}
In the introduction, we motivated the use of weak labels by raising the issues of limited strong supervision, complementarity between strong and weak supervision, the need to use more complex models to solve challenging problems and the heterogeneity in the data population. In this section we will present additional experiments to analyze each one of these factors.

\subsubsection{Effect of more strongly-labeled training samples}
\begin{figure}
    \centering
    \includegraphics[width=0.4\textwidth]{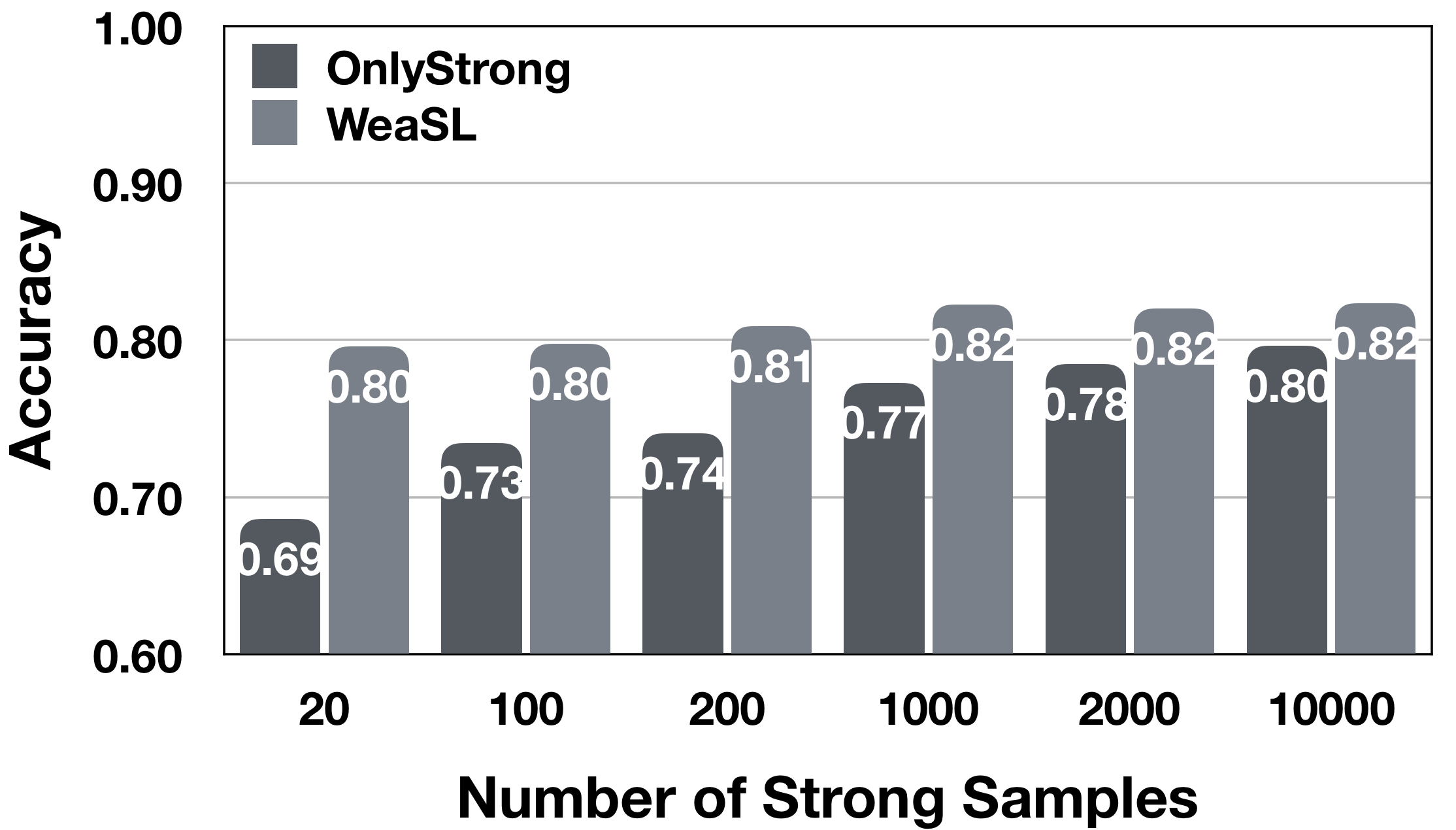}
    \caption{Gain with increasing number of strongly-labeled samples: The x-axis shows the number of strongly labeled samples provided to the algorithms.}
    \label{figLabaledSamplesExpPlotFires}
    \vspace{-0.5cm}
\end{figure}

\emph{WeaSL} tries to leverage the abundant weakly labeled data to improve the performance of a model that just uses strongly labeled data, typically available in small numbers. As the number of strongly-labeled instances increases, the gain obtained by using the weakly-labeled instances reduces. We demonstrate this empirically by increasing the number of strongly labeled instances in data set D2 from $10$ up to $10000$. From the results in figure \ref{figLabaledSamplesExpPlotFires}, we observe that the $accuracy$ values for all methods increase as the number of strongly labeled samples are increased. Finally, given enough number of samples, \emph{OnlyStrong} catches up to \emph{WeaSL} in terms of its accuracy value.

\subsubsection{Complementarity between strong and weak supervision}
\begin{figure}
    \centering
    \includegraphics[width=0.4\textwidth]{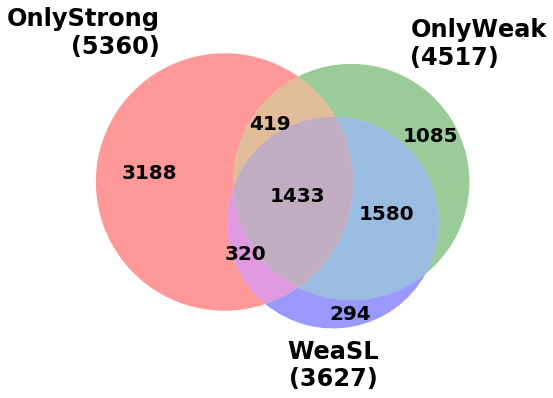}
    \caption{Venn diagram of errors made on D1 by different models. \emph{WeaSL} leverages the complementarity between models trained with only strong and only weak supervision separately.} %\emph{WeaSL} significantly reduces the errors relative to both \emph{OnlyStrong} and \emph{OnlyWeak}, while not adding a significant number of new errors.}
    \label{figComplementarity}
    \vspace{-0.5cm}
\end{figure}

As explained in the introduction, by making use of strong and weak supervision together, \emph{WeaSL} attempts to overcome the limitations of both. Here, we try to empirically demonstrate the complementarity between strong and weak supervision and its benefit to \emph{WeaSL}. Figure \ref{figComplementarity} shows a Venn diagram of the errors made by models trained with strong supervision (\emph{OnlyStrong}), weak supervision (\emph{OnlyWeak}) and both (\emph{WeaSL}) on the data set D1. Firstly, we can see that the total number of errors made by \emph{WeaSL} is less than \emph{OnlyStrong} and \emph{OnlyWeak}. Secondly, a significant number of errors made individually by either \emph{OnlyStrong} or \emph{OnlyWeak} are not made by \emph{WeaSL}. In particular, $67\%$ of the errors made by \emph{OnlyStrong} are not made by \emph{WeaSL}. Similarly, $33\%$ of the errors made by \emph{OnlyWeak} are not made by \emph{WeaSL}. Thirdly and perhaps, most importantly, only a very small fraction of the errors made by \emph{WeaSL} are new i.e not made by either \emph{OnlyStrong} or \emph{OnlyWeak}. Thus \emph{WeaSL} significantly reduces the errors relative to both \emph{OnlyStrong} and \emph{OnlyWeak}, while not adding a significant number of new errors.

\subsubsection{Increasing model complexity with weak labels}
\begin{figure}
    \centering
    \includegraphics[width=0.35\textwidth]{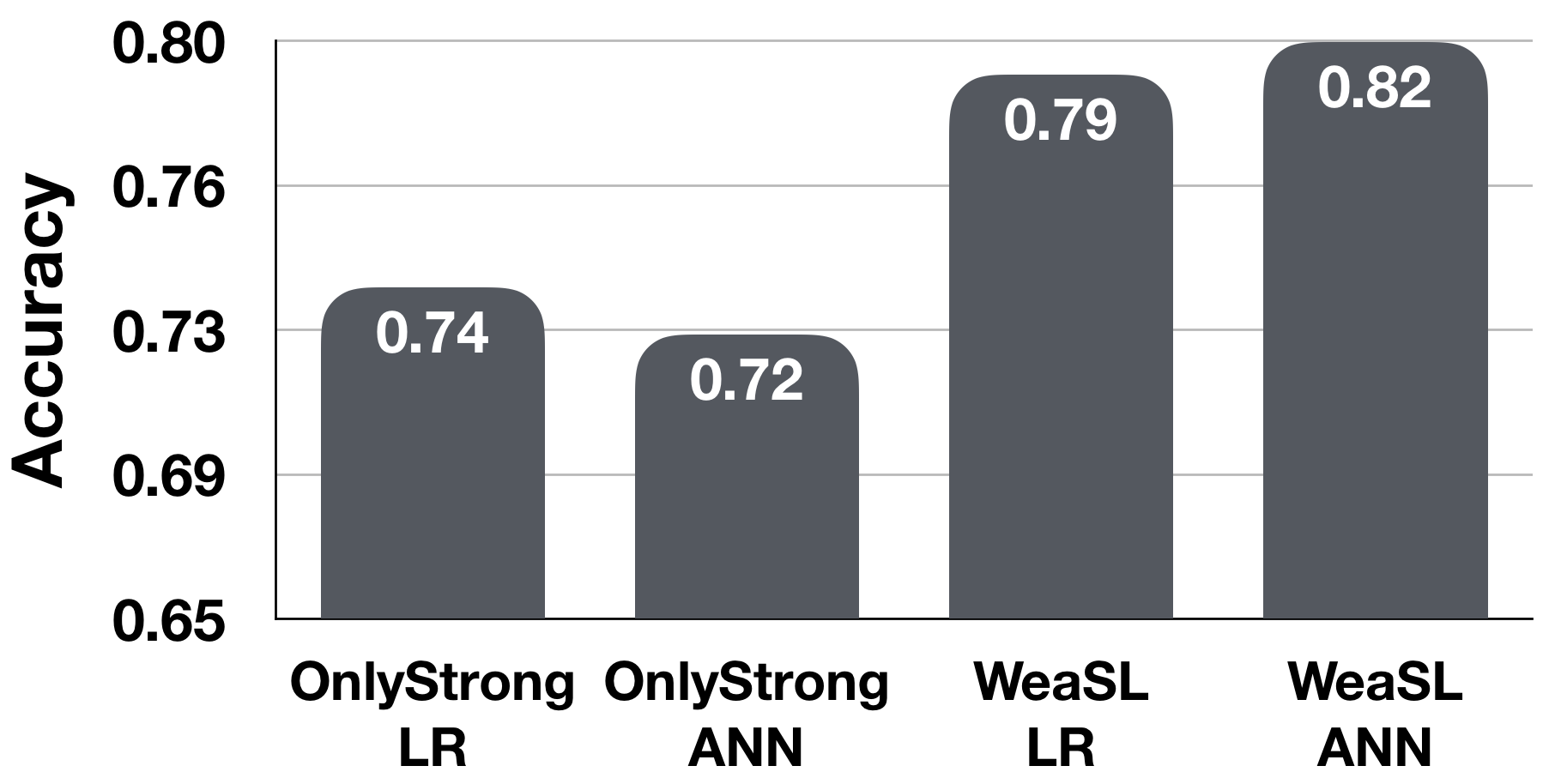}
    \caption{Using weak labels enables training more complex models.}% If we rely on only strong-labeled samples, complex models might tend to overfit and generalize poorly. However, with the use of weak labels, we get the benefit of their increased complexity.}
    \label{figModelComplexity}
    \vspace{-0.5cm}
\end{figure}

The second key motivation for using weak labels is that for many applications, simpler models do not suffice and we would like to use models with more parameters like deep neural networks. However, the more complex the model, the more samples needed to train it. Here again, weak labels can be used to improve generalization performance of complex models that would otherwise have poor generalization performance due to limited strongly-labeled samples. To demonstrate this, we train a logistic regression model and a artificial neural network (trained using dropout \cite{dropout} regularization) with two hidden layers on the data set D2. The ANN is a more complex model, so it is expected to overfit on the training data if we use only strong labeled samples. However, once we use weak labeled samples as well, it outperforms the logistic regression model, as shown in figure \ref{figModelComplexity}.

\subsubsection{Mitigating heterogeneity with weak labels}
In the presence of heterogeneity in the data population, one needs samples from every homogeneous variety of instances in the data for the classifier to have good generalization performance. However, the generalizability of a model trained with only strong instances improves when weak supervision added to its training. This can be inferred to some extent by the results in section \ref{baselineComp}. To demonstrate this effect further, we test the performance of a model trained on one data set as tested on a completely different data set, both relating to fires. The results in table \ref{tableMitigatingHeterogeneity} show that the model trained on the source data set does not do well on the target data set, when trained just with strong supervision, however, adding weak supervision significantly improves the performance, thus demonstrating the generalizability of the model to heterogeneous regions.

\begin{table}[h]
    \centering
    \caption{Mitigating heterogeneity with weak labels: Table reports accuracy as models trained on a source data set are applied on a different target data set.}
    \begin{tabular}{|p{1.3cm}|p{1.3cm}|p{1.3cm}|p{1.4cm}| p{1.6cm}|}
        \hline
        Source & Target & OnlyStrong &  OnlyWeak &  WeaSL  \\ \hline
        D2 & D1 & 0.565 & 0.604 & \textbf{0.682} \\ \hline
        D1 & D2 & 0.637 & 0.532 & \textbf{0.684} \\ \hline
    \end{tabular}
    \label{tableMitigatingHeterogeneity}
    \vspace{-0.5cm}
\end{table}

\subsection{Sensitivity Experiments}
In this section, we test the sensitivity of the proposed method to changes in different aspects of the problem setting. We consider 3 different factors of the problem setting and measure the performance of our methods (and some baselines for comparison) as these factors are changed. The 3 factors that we consider are - skew, purity of instance-labels relative to group labels.

\subsubsection{Skew}
\begin{figure}
    \centering
    \includegraphics[width=0.43\textwidth]{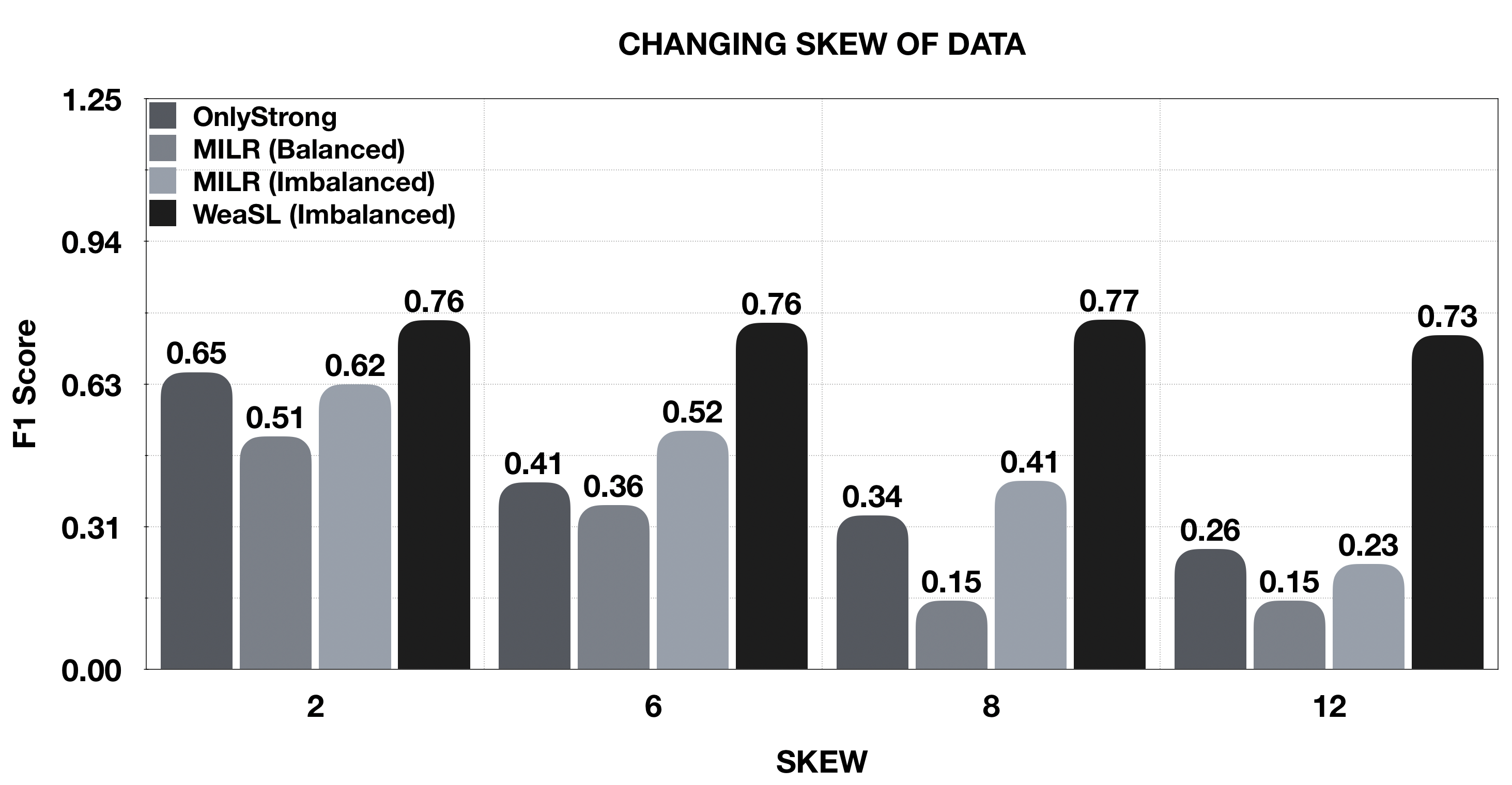}
    \caption{Tolerance to skew: The x-axis shows the skew, defined as the ratio of negatives (label 0) to positives (label 1) instances. Notice how the relative benefit of using \emph{Imbalanced WeaSL} increases as the skew in the data is increased.}\label{skewExpPlot}
    \vspace{-0.5cm}
\end{figure}

One of the key aspect of \emph{WeaSL} is its ability to model imbalance in classes and thus not be biased to perform well on the majority class and avoid false-positives. In this experiment, we consider balanced data set D2 and sample the instances in the positive class so that the skew in the new data set progressively increases from 1:2 initially to 1:12 in the end. The strongly-labeled data set is still balanced, while the weakly-labeled training set, validation set and test set are skewed. Note that the groups are not broken during this process, they are sampled so that the desired skew in instances is maintained. Figure \ref{skewExpPlot} shows the F-measure for each setting of class skew. The x-axis shows the skew in the data defined as the ratio of negatives (label 0) to positives (label 1). Notice how the benefit of using \emph{WeaSL} over other methods increases as the skew in the data is increased.

\subsubsection{Purity of groups} \label{PurityExpSection}
\begin{figure}
    \centering
    \begin{subfigure}{0.3\textwidth}
        \centering
        \includegraphics[width=\textwidth]{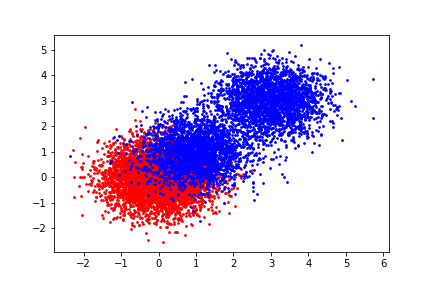}
        \caption{Synthetic data distribution. Red denote positives, blue denote negatives}
        \label{figPurityExpDist}
    \end{subfigure}

    \begin{subfigure}{0.5\textwidth}
        \centering
        \includegraphics[width=\textwidth]{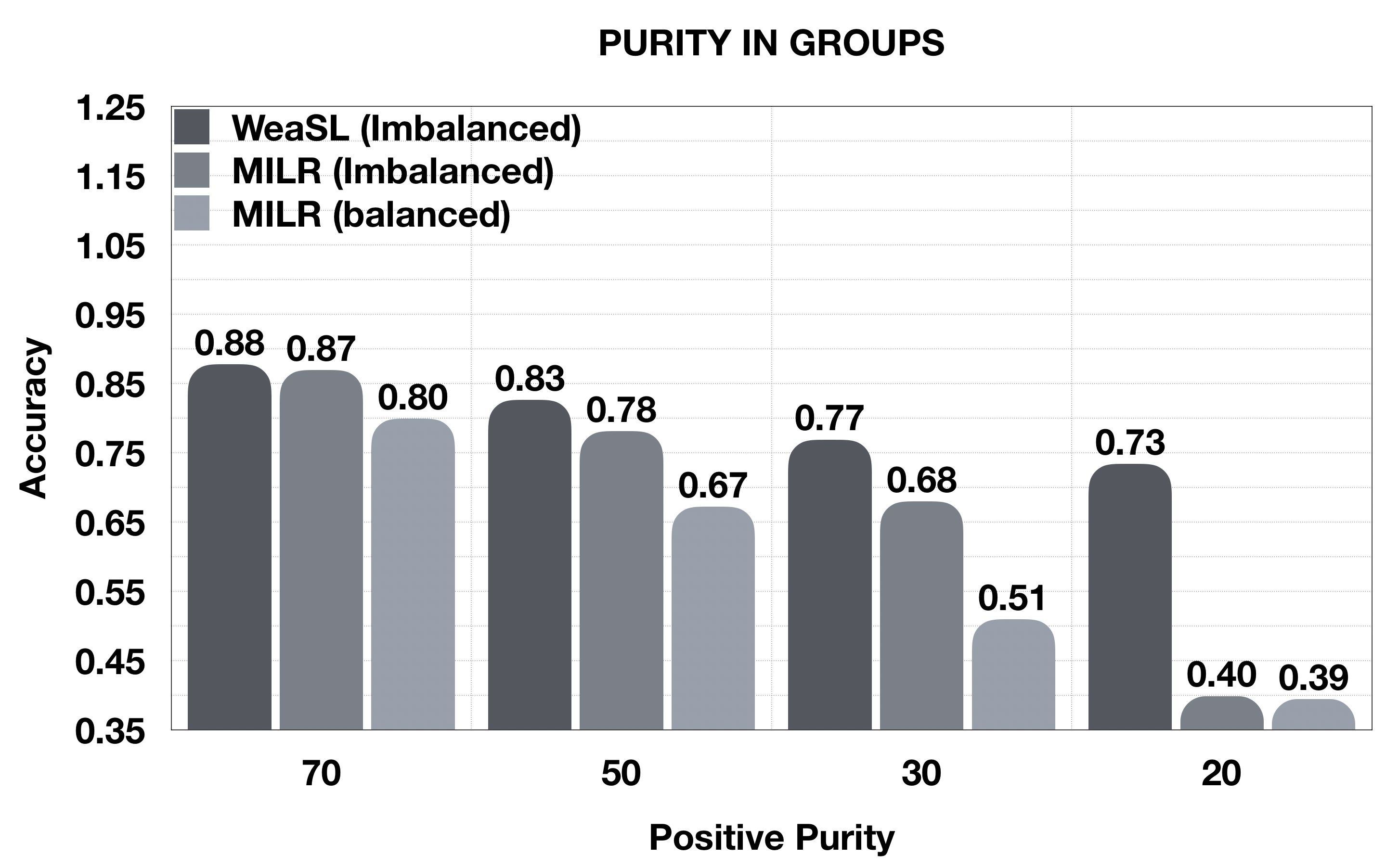}
        \caption{Effect of varying purity of groups in weakly-labeled data}
        \label{figPurityExpPlot}
    \end{subfigure}
\caption{Tolerance to purity of groups in weakly-supervised set: \ref{figPurityExpDist} shows the distribution from which instances in each class are sampled from. \ref{figPurityExpPlot} shows the performance of different methods as group purity $f$ is varied. X-axis shows values of $f$ while the height of the bars show the performance in F-score.} %\emph{Imbalanced WeaSL} is much more tolerant to decreases in purity of the groups than other methods, giving reasonable performance even at low purity levels. Moreover, the benefit of using \emph{Imbalanced WeaSL} relative to other methods increases with decreasing purity, since the explicit modeling of noise and rarity becomes more relevant.}
\label{figPurityExp}
% \vspace{-1cm}
\end{figure}

In this experiment, we change the purity of groups in the weakly-labeled data, where the purity of a group is defined as the number of instances in the group that have the same label as the group-label. For this experiment, we create a synthetic data set with 2-dimensional features, where the positive instances are sampled from 2 Gaussians with means (1,1), (3,3) respectively and standard deviation 1 in each dimension while the negative instances are sampled from a Gaussian distribution with mean (0,0) and standard deviation 1 in each dimension. Figure \ref{figPurityExpDist} shows the scatter plot for the instances, where the positive instances are colored red while the negative instances are colored blue. For the weakly-labeled data set, groups of size $20$ are created, where the negatively-labeled groups are $100\%$ pure i.e they are made of entirely negative instances. The positive groups, on the other hand, have a certain fraction $f$ of their 20 instances as positive, while $(1-f)$ fraction instances in the positive groups are negative.  The weakly-labeled data has equal number of positive and negative groups i.e the group skew is 1. Note that in this setting, the instance skew i.e the ratio of negative to positive instances in the weakly-labeled set is $(2/f -1)$. Thus, as the purity of the positive bags, encoded by the fraction $f$ decreases, the skew in the data set increases. Therefore, if one were to just optimize group-level predictions, one would be totally oblivious to the underlying skew in the instances. For each value of group purity $f$, we learn the algorithms \emph{Balanced MIL}, \emph{Imbalanced MIL}, \emph{Imbalanced WeaSL} and \emph{OnlyStrong} and test them on a data set with the same instance-level skew i.e $(2/f -1)$. Figure \ref{figPurityExpPlot} shows the relative performance of the different algorithms as the purity of the groups is reduced. Note than the \emph{Imbalanced WeaSL} is much more tolerant to decreases in purity of the groups than other methods, giving reasonable performance even at low purity levels. Moreover, the benefit of using \emph{Imbalanced WeaSL} relative to other methods increases with decreasing purity, since the explicit modeling of noise and rarity becomes more relevant.

\section{Conclusion}
The paradigm of weakly-supervised learning \cite{zhouWLIntro} has shown great promise in mitigating the problem of not having adequate labeled samples for training predictive models. In this paper, we proposed a new problem setting where binary labels,  available for groups of instances, can be used as weak labels for regularizing a model trained for an instance-level binary classification problem. We modeled the group-labels as Class Conditional Noisy (CCN) labels for instances and proposed a learning algorithm \emph{WeaSL} that regularizes the performance over weakly-labeled instances by optimizing the desired metric i.e accuracy for balanced class problems and G-measure for imbalanced class problems. Our experiments over real-world applications of land cover monitoring problems show the utility of the proposed approach and its superiority over other baseline techniques to solve this problem. We also demonstrate that \emph{WeaSL} is robust to changes in rarity of the classes, purity of instances within groups and number of available strongly-labeled instances. Moreover, we show that \emph{WeaSL} demonstrably leverages complementarity between the two sources of supervision - strong and weak, enables the use of more complex models and helps mitigate data heterogeneity, which are some of the key factors that motivated the use of group-level weak labels. Further research includes relaxing the CCN assumption for modeling group-level labels and exploring other kinds of weak labels for binary classification problems.
% \vspace{-1cm}

\bibliographystyle{ACM-Reference-Format}
\bibliography{main}

\end{document}